\documentclass{article}
\usepackage{spconf,amsmath,graphicx}
\usepackage{amsmath,amssymb,amsfonts}
\usepackage{algorithm}
\usepackage{algorithmic}

\usepackage{graphicx}
\usepackage{textcomp}
\usepackage{xcolor}
\usepackage{subfigure}
\usepackage{multirow}
\usepackage{mathrsfs}

\title{Extracting the Brain-like Representation by an Improved Self-Organizing Map for Image Classification}
\name{
    Jiahong Zhang,\textsuperscript{\rm 1,2}
    Lihong Cao, \textsuperscript{\rm 1,2 } 
    Moning Zhang, \textsuperscript{\rm 1,2}
    Wenlong Fu\textsuperscript{\rm 1,2}
    \thanks{The corresponding author is Lihong Cao (lihong.cao@cuc.edu.cn).}
}
\address{
    \textsuperscript{\rm 1} State Key Laboratory of Media Convergence and
Communication, Communication University of China,\\
    \textsuperscript{\rm 2} Neuroscience and Intelligent Media Institute, Communication University of China,\\Beijing,China
}

\begin{document} \normalsize
%
\maketitle
\begin{abstract}
Backpropagation-based supervised learning has achieved great success in computer vision tasks. However, its biological plausibility is always controversial. Recently, the bio-inspired Hebbian learning rule (HLR) has received extensive attention. Self-Organizing Map (SOM) uses the competitive HLR to establish connections between neurons, obtaining visual features in an unsupervised way. Although the representation of SOM neurons shows some brain-like characteristics, it is still quite different from the neuron representation in the human visual cortex. This paper proposes an improved SOM with multi-winner, multi-code, and local receptive field, named mlSOM. We observe that the neuron representation of mlSOM is similar to the human visual cortex. Furthermore, mlSOM shows a sparse distributed representation of objects, which has also been found in the human inferior temporal area. In addition, experiments show that mlSOM achieves better classification accuracy than the original SOM and other state-of-the-art HLR-based methods. The code is accessible at https://github.com/JiaHongZ/mlSOM.
\end{abstract}

\section{Introduction}
Backpropagation-based supervised learning has been extensively studied in recent years. For image classification, the adoption of backpropagation enables convolutional neural networks (CNNs) to extract features effectively \cite{AlexNet,VGG,ResNet}, thus improving classification performance. However, the biological plausibility of the backpropagation is always controversial \cite{2000Computational}.

\begin{figure}[t]
	\centering	
    \centering\includegraphics[width=7cm,height=7cm]{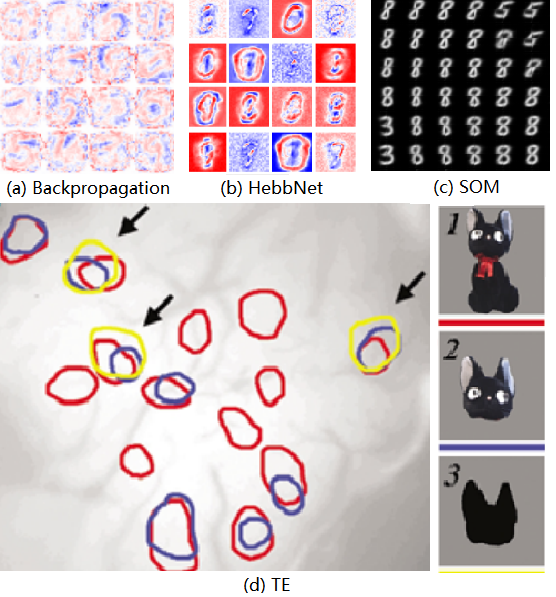}

	\caption{Visualization of the neuron representations:  (a) HebbNet, (b) Backpropagation, (c) SOM for handwritten digits. (d) TE for objects (Image Source: \cite{ITPart}). Colored bars in the top (red), middle (green) and bottom graphs in (d) are penetration sites inside the active spots of the stimulus.}
	\label{fig1}
\end{figure}

\begin{figure*}[!t]
	\centerline{\includegraphics[width=15cm]{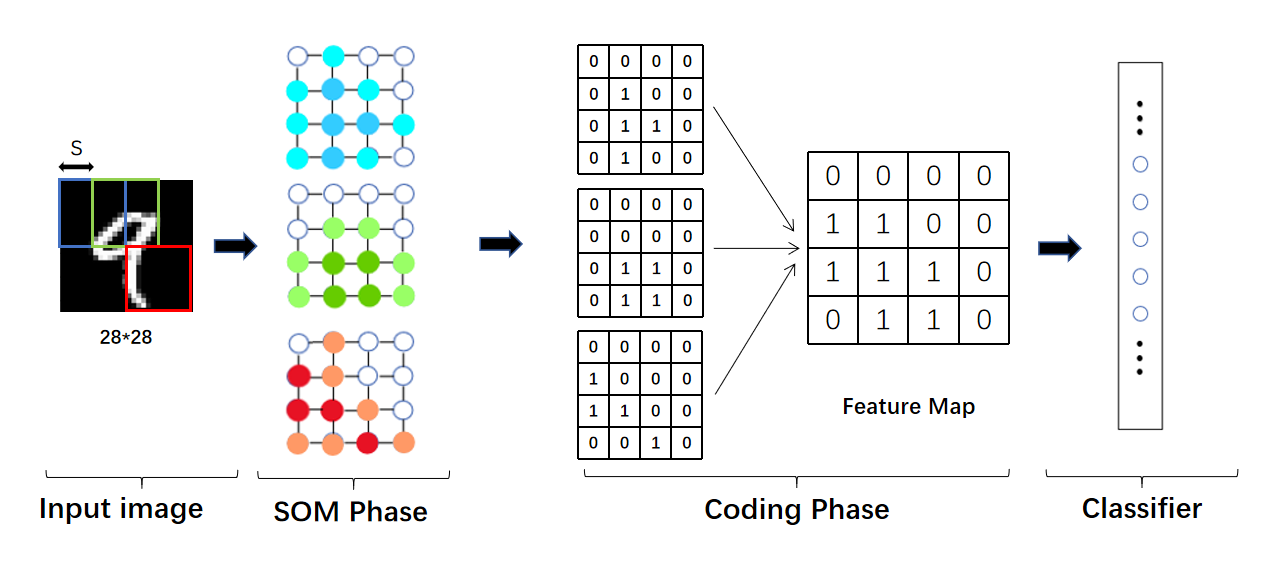}}
	\caption{ The architecture of mlSOM. For a given input image, mlSOM first divide it into patches through a sliding window. Then for each patch, its WNs in the hidden layer are calculated in the SOM phase. In the Coding phase, the neuron states of the hidden layer will be coded to a feature map, in which $1$ denotes the WN. The feature map is then send to the linear classifier to obtain the classification result.
	}
	\label{fig2}
\end{figure*}

The Hebbian learning rule (HLR) is a biologically plausible unsupervised learning mechanism and has been proposed for a long time \cite{allport1985distributed,klopf1972brain}, which suggests that: "Neurons that fire together wire together." In a broad sense, the HLR refers to a family of methods based on the idea of Hebbian. Unfortunately, vanilla HLR does not guarantee high performance for image classification. Recently, several methods have been proposed to improve the classification accuracy for HLR  \cite{hopfieldhebb,HebbNet,amato2019hebbian}. Self-Organizing Map (SOM) uses the Winner-Takes-All competition HLR  to establish connections between neurons \cite{rumelhart1985feature, 1992Neural}, which achieves high classification performance. However, they failed to obtain brain-like neuron representation. In human visual cortex, the representation of an object presents a topological structure \cite{ITPart}. For example, Fig.~\ref{fig1} (d) shows the representation of a complex object and its parts in the anterior part of the IT cortex (architectonically defined as area TE). By comparison, it can be found that the neuron representations of existing HLR methods (Fig.~\ref{fig1} (b) and (c)) and the backpropagation method (Fig.~\ref{fig1} (a)) lack the object parts.

This paper proposes an improved SOM, mlSOM. Compared with the original SOM, three modifications are made in mlSOM: from global receptive field (GRF) to local receptive field (LRF), from single-winner to multi-winner, and from single-code to multi-code. 
The main contributions of this work contain at least three key advantages. Firstly, we propose to improve the representation of SOM from a brain-like perspective, which may contribute to a promising future research direction for SOM. Secondly, we improve the original SOM in three ways inspired by the human brain and demonstrate their effectiveness for classification. Thirdly, the proposed mlSOM shows brain-like representation and gets high classification performance compared with other state-of-the-art Hebbian learning-based methods.

\begin{table}[t]\footnotesize 
\renewcommand\arraystretch{1}
	\caption{The hyper-parameters of mlSOM.}
	\begin{center}
		\begin{tabular}{c|c|c|c|c|c|c|c}
			\hline
			\textbf{Datasets} & \multicolumn{4}{c}\textbf{Hyper-parameters}\\ 	\hline	
			
			& \textit{hidden neurons}
			& \textit{w}
			& \textit{s}
			& \textit{$n$}
			& \textit{$\sigma$} & \textbf{$k$} & \textbf{$lr$}\\
			\hline
			\multirow{1}{*}{MNIST}& $44\times44$ & $14\times14$ & 7 & 5 & 2 & 20 & 0.3\\
			\hline
			\multirow{1}{*}{CIFAR-10} 
			& $44\times44$ & $16\times16$ & 4 & 5 & 2 & 100 & 0.3\\
			\hline
		\end{tabular}
		\label{tab1}
	\end{center}
\end{table}

\section{Related Work}
Self-Organizing Map (SOM) is a kind of HLR-based neural network. For image classification, current studies for SOM mainly focus on improving classification accuracy. Supervised SOM was proposed in \cite{supervisedSOM3} and got good classification results. Some work presented that deep SOM would get higher classification performance than the single-layer SOM \cite{dsom1, dsom2}. Combining SOM and CNN to obtain both accuracy and biological plausibility has also attracted widespread interest \cite{somcnn1,somcnn2,somcnn3,somcnn4}. This paper presents a new idea to improve SOM from the neuron representation perspective.

\begin{algorithm}[!h] \small
    \caption{Learning algorithm for mlSOM}
    \begin{algorithmic}[1]
    \REQUIRE ~~\\ 
    Training set of images and labels ($\mathscr{X}, Y$) \\
    \ENSURE 
    Trained mlSOM 
    \STATE initialize the model parameter $W$ with the standard normal distribution and hyper-parameters in Table \ref{tab1}\\
	{lr, epochs $\leftarrow$ initialize the learning rate, 
	epochs} \\
    
    \STATE \% SOM Phase (unsupervised)
    \FOR{epo $\in$ epochs}
       \FOR{$x \in \mathscr{X}$}
        \STATE $ x_p \leftarrow Sliding(x)$ \% image patches obtained by the sliding window;
        \FOR{$x_{p_i} \in x_p$}
        \STATE \% distance of $x_{p_i}$ and neurons in mlSOM
        \FOR {$W_j \in W$}
        \STATE $d_{ij} \leftarrow  \Vert x_{p_i} - W_j \Vert$
        \ENDFOR
        \STATE sort $d_{ij}$ in ascending order;
        \STATE WNs $\leftarrow$ the top $n$ $W$;
        
        \STATE $[\left(X_{WN},Y_{WN} \right)] \leftarrow$ the coordinate of the first $n$ WNs;
        
        \STATE \% update learning rate
        \STATE $lr_{epo} \leftarrow lr \times \left( 1- \frac{epo}{epochs}\right)$;

        \FOR {$WN_i \in WNs$}
        \FOR {$W_j \in W$}
        \STATE \% compute updating neuron vector decay value;
        \STATE $decay = e^{\frac{{\Vert \left( X_{WN_i},Y_{WN_i} \right) - \left( X_{W_j},Y_{W_j} \right) \Vert}_2 }{2\sigma^2}}$ ;
		\STATE $\Delta W = lr_{epo} \times decay \times ( W_i - WN$);
		\STATE $W_i \leftarrow W_i + \Delta W$ ;
        \ENDFOR
        \ENDFOR
        \ENDFOR
        \ENDFOR
        \ENDFOR    
        \STATE \% Coding phase (supervised)
        \FOR{$epo \in epochs$}
            \FOR{$x \in \mathscr{X}$}     
            \STATE $ x_p \leftarrow Sliding(x)$;
            \FOR{$x_{p_i} \in x_p$}
            \STATE $G_p \leftarrow$ binary 2D grid with $k$ WNs of the hidden layer for $x_p$;
            \STATE $G_{sum} = Binary(\sum{G_p})$;
    		\STATE feature map $\leftarrow$ $G_{sum}$;
    		\STATE \% prediction of the classifier
    		\STATE $pre = f(feature map)$;
    		\STATE minimize $L(pre, Y)$;
            \ENDFOR
            \ENDFOR
        \ENDFOR
    \end{algorithmic}
    \label{al2}
\end{algorithm}

\section{Proposed method}
The proposed mlSOM is based on the unsupervised SOM algorithm mentioned in \cite{dsom1}. We first revisit it.
\subsection{The original SOM}
SOM is a classic unsupervised learning algorithm using the "winner-take-all" learning rule, which gets a non-linear projection of high-dimensional data over a small space. Each neuron in SOM consists of a trainable vector. SOM computes Euclidean distances between the input pattern and each neuron. The neuron that has the least distance is the winner neuron (WN). WN and its neighbors will be updated to be closer to the input pattern during training. The update value decreases as the distance between neurons and WN increases.

\subsection{mlSOM}
Fig.~\ref{fig2} shows the architecture of mlSOM whose hyper-parameters are illustrated in Table~\ref{tab1}. mlSOM is based on SOM, and the three modifications in mlSOM are as follows.

\textbf{1) From GRF to LRF.} SOM computes the distance between the whole input image and hidden layer neurons, leading to huge neuron vectors. Inspired by the human eye movement, we propose to use LRF, which can be realized by a sliding window. As shown in Fig.~\ref{fig2}, the size of the input image is $(H,W)$ and the window size is set to $(w, w)$ with stride $(s,s)$. The size of the neuron vector in mlSOM is $w^2$, which is $(\frac{w^2}{H \times W})$ of the original SOM.


\textbf{2) From single-winner to multi-winner.} The original SOM uses the "winner-take-all" learning rule. It computes Euclidean distances between the input image and neurons in the hidden layer and chooses one WN. However, population coding widely exists in the human visual cortex \cite{golledge2003correlations}. It motivates us to change the single-winner to multi-winner. As shown in Fig.~\ref{fig2} SOM Phase, for every image patch, the hidden layer of mlSOM obtains one 2D grid with $n$ winners. For every winner, the vector-updating algorithm is similar to the original SOM. 

\textbf{3) From single-code to multi-code.} The original SOM and some deep SOMs use the 2D grid with a single WN as the classification feature map. Also inspired by the population coding, mlSOM generates the feature map with multiple WNs, as shown in Fig.~\ref{fig2} Coding Phase.
Specifically, mlSOM uses neurons with the first $k$ WNs to achieve the multi-code. Here, $k$ can be different from the multi-winner $n$. In Fig.~\ref{fig2} Coding phase, the 2D grids of input image patches are transformed to the corresponding binary matrixs, in which $1$ denotes the WN. These matrices are summed together and binarized as the feature map of the input image.

mlSOM is an unsupervised learning algorithm. We trained a linear classifier to classify images using their mlSOM feature maps.

\begin{figure*}[!t]
	\centering	
	\subfigure[representations]{
		\centering\includegraphics[width=6.5cm]{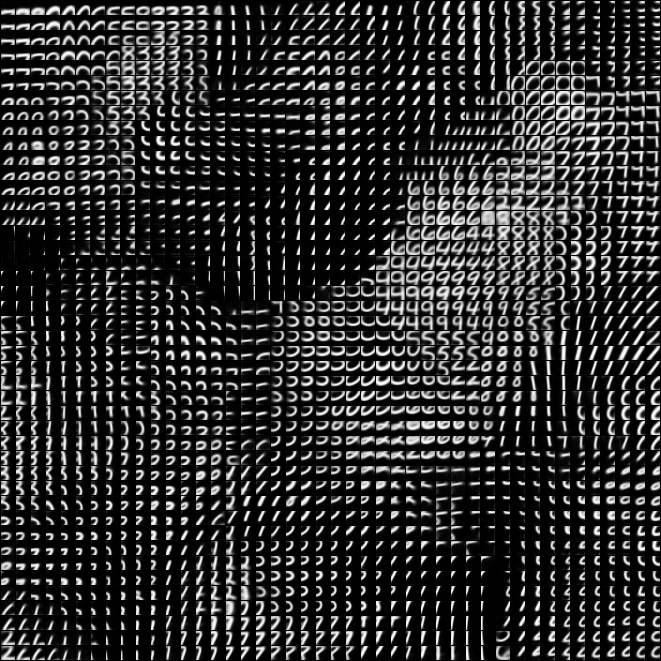}
	}
	\subfigure[coding]{
		\centering\includegraphics[width=9cm]{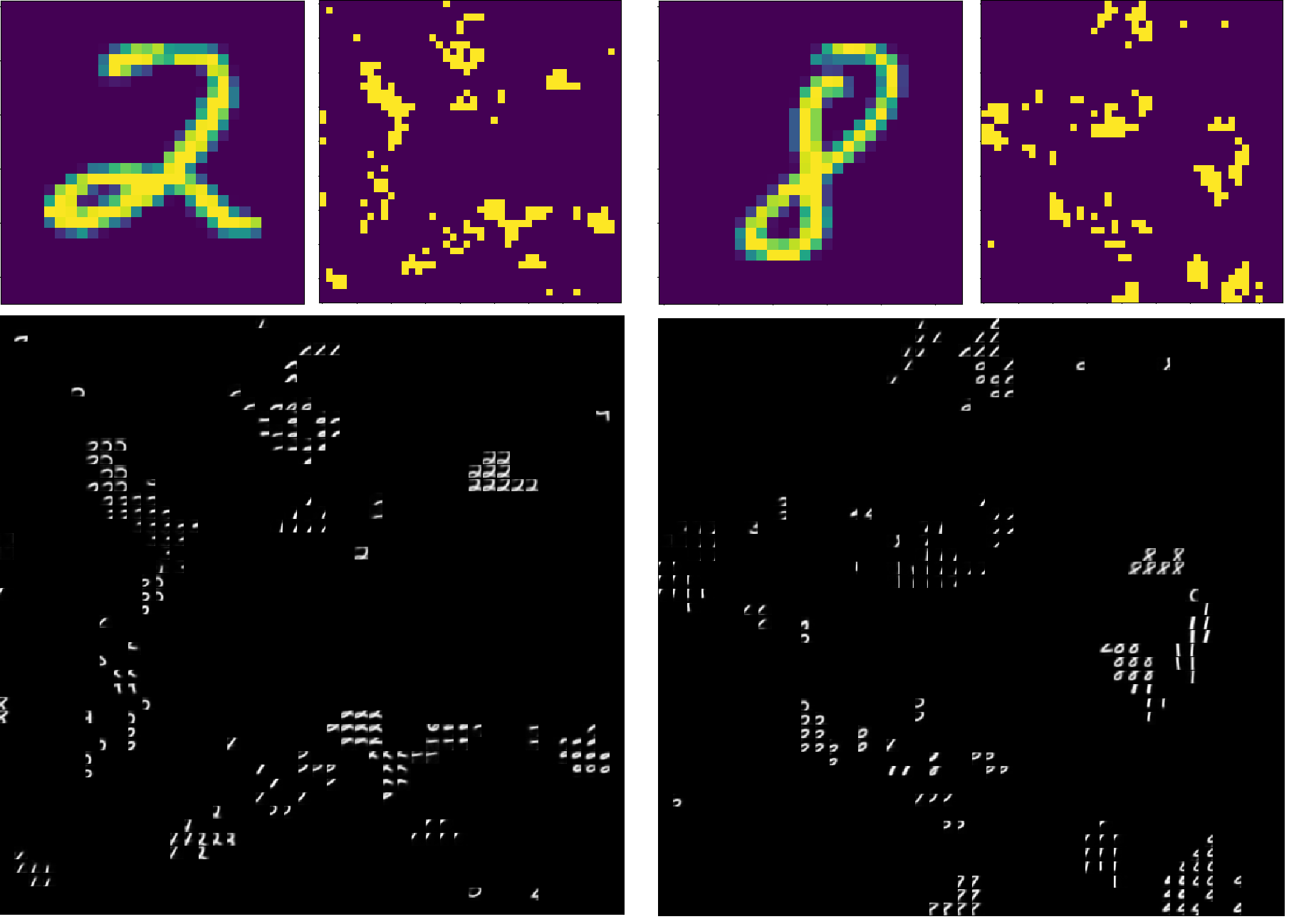}
	}
	\caption{(a) Neuron representations learnt by the mlSOM after training on MNIST. (b) Visualization of the feature maps and neuron representations of the specific input images. The top row shows the input images and their feature maps. Yellow dots in feature maps represent the WNs. The bottom row shows the neuron vectors of the WNs.}
	\label{pa}
	\vspace{-15pt}
\end{figure*}

\subsection{Training method}
The training method of mlSOM is shown in Algorithm \ref{al2}. 
This training process can be divided into two phases. In the SOM phase, an input image is firstly divided into patches by the sliding window. Then, Euclidean distances of the image patches and the hidden layer neurons are computed. The top $n$ neurons with minimum distance will be selected as WNs. For each neuron of the WNs, the algorithm updating the neuron vectors in mlSOM is similar to the original SOM. 

When the SOM phase is finished, we get a trained hidden layer. During the coding phase, for each input image patch, a corresponding 2D grid is obtained from the hidden layer of mlSOM. We convert this 2D grid into a binary coding matrix, in which $1$ denotes the WN. Here, the number of $1$ in the coding matrix is set to $k$. Then, we obtain the sum of all the patch grids and binarize it to get the feature map representing a specific object. 
To verify the classification ability of the feature map, we train a linear classifier with the help of error backpropagation. The training object is minimizing the cross-entropy loss:
\begin{equation}
	L(x) = \sum_{c=1}^{N}ylog(f(x)),
\end{equation}
where x denotes the input image, $f(\cdot)$ denotes the classifier, and y denotes its label.

\section{Experiments}

\subsection{Experiment results} \label{results}
We use two datasets MNIST\cite{mnist}, CIFAR-10\cite{cifar10} to evaluate the proposed mlSOM. This section compares mlSOM to some popular Hebbian learning-based methods. We choose the methods with a single hidden layer for a fair comparison. As shown in Table \ref{mnist}, mlSOM performs significantly better classification results than that of HebbNet and SOM, achieving 96.79\% test accuracy on MNIST. According to the ablation experiments results in Table \ref{mnist}, the three modifications, LRF, multi-winner, and multi-code, effectively contribute to better performance. Table \ref{cifar} shows the results on CIFAR-10. It can be observed that mlSOM can achieve competitive results with the state-of-the-art methods. Furthermore, mlSOM trades off the classification accuracy and the convergence speed.

\begin{table} \small 
	\renewcommand\arraystretch{1} 
	\caption{Classification accuracy results and ablation experiments results of mlSOM on the MNIST dataset.}
	\begin{center}  
	\begin{tabular}{ll} 
		\hline                      
		\textbf{Method} & Test Accuracy  \\  
		\hline  
		Vanilla Hebbian  &  10.28\\ 
		HebbNet\cite{HebbNet}  &  93.25\\ 
		SOM  & 93.07\\
		DSOM \cite{dsom1} & 96.17 \\
		\hline 
		SOM+mult-winner  & 95.40\\
		SOM+mult-winner+LRF  & 96.72\\
		\textbf{mlSOM}  & \textbf{96.79}\\      
		\hline  
	\end{tabular}  
\end{center}
\vspace{-25pt}
\label{mnist}
\end{table}

\subsection{Neuron representations of mlSOM}
According to Fig.~\ref{fig1} and Fig.~\ref{pa}, the neuron representation obtained by mlSOM exhibits brain-like characteristics which are similar to the neuron activity detected in the human TE \cite{ITPart}. 
Specifically, we found that neurons in mlSOM respond to the whole and part of the object. Furthermore, mlSOM presents a brain-like distributed coding method. Fig.~\ref{pa} (b) shows the feature map and neuron representations in mlSOM for digits two and eight. Taking the digit two as an example, its representations in mlSOM consist of the arcs, slashes, object features, and corresponding parts. These representations are observed in the human visual cortex \cite{kandel2000principles}. It is worth noting that neurons representing slashes are also detected in digit eight. This evidence suggests that neurons in mlSOM can respond to feature combinations of different objects, contributing to a larger encoding capacity. It is very similar to the sparse distributed representation in human IT \cite{rolls1995sparseness,rolls1997information,franco2007neuronal}.

\begin{table}  \small 
	\renewcommand\arraystretch{1} 
	\caption{Classification accuracy results on the CIFAR-10 dataset.}
	\begin{center}  
		\begin{tabular}{llll} 
			\hline                      
			\textbf{Method} & Train Acc & Test Acc & Epochs \\  
			\hline  
			Vanilla Hebbian  &   11.56 &  15.23 &  200\\ 
			BackProp  &  39.89 &  41.28 & 200\\ 
			Krotov et al. \cite{hopfieldhebb} & \textbf{55.05} & \textbf{50.75} & 1500\\
			Amato et al. \cite{amato2019hebbian} & - & 41.78 & \textbf{20}\\
			HebbNet \cite{HebbNet} & 43.13 & 45.69 & 200\\
			\textbf{mlSOM}  & 51.82 & 43.65 & 200 \\      
			\hline  
		\end{tabular}  
	\end{center}
	\vspace{-20pt}
	\label{cifar}
\end{table}

\section{Conclusion}
\vspace{-5pt}
This paper proposes an improved SOM method, mlSOM, which achieves better classification accuracy than other Hebbian learning-based methods on MNIST and competitive accuracy on CIFAR-10. mlSOM makes improvements based on the brain-inspired designs, including LRF, multi-winner, and multi-code. Ablation experiments show the effectiveness of these three modifications. The most significant contribution of mlSOM is that it exhibits brain-like neuronal representations and coding. Our research may inspire the design of visual cortex computing model and provide a novel direction for SOM research.
\vspace{-10pt}
\section{Acknowledgment}
\vspace{-5pt}
This paper is supported by supported by the STI 2030—Major Projects (grant No. 2021ZD0200300) and the National Natural Science Foundation of China (grant No. 62176241).

\label{sec:refs} \small

\bibliographystyle{IEEEbib}
\bibliography{strings,refs}

\begin{thebibliography}{10}

\bibitem{AlexNet}
Alex Krizhevsky, I.~Sutskever, and G.~Hinton,
\newblock ``Imagenet classification with deep convolutional neural networks,''
\newblock {\em Advances in neural information processing systems}, vol. 25, no.
  2, 2012.

\bibitem{VGG}
Karen Simonyan and Andrew Zisserman,
\newblock ``Very deep convolutional networks for large-scale image
  recognition,''
\newblock {\em arXiv preprint arXiv:1409.1556}, 2014.

\bibitem{ResNet}
Kaiming He, Xiangyu Zhang, Shaoqing Ren, and Jian Sun,
\newblock ``Deep residual learning for image recognition,''
\newblock in {\em 2016 IEEE Conference on Computer Vision and Pattern
  Recognition (CVPR)}, 2016, pp. 770--778.

\bibitem{2000Computational}
R.~C. O'Reilly and Y.~Munakata,
\newblock ``Computational explorations in cognitive neuroscience: Understanding
  the mind by simulating the brain,''
\newblock {\em MIT Press}, 2000.

\bibitem{ITPart}
Kazushige Tsunoda, Yukako Yamane, Makoto Nishizaki, and Manabu Tanifuji,
\newblock ``Complex objects are represented in macaque inferotemporal cortex by
  the combination of feature columns,''
\newblock {\em Nature neuroscience}, vol. 4, no. 8, pp. 832--838, 2001.

\bibitem{allport1985distributed}
D~Allport,
\newblock ``Distributed memory, modular systems and dysphasia| bibsonomy,''
\newblock {\em Current Perspectives in Dysphasia}, 1985.

\bibitem{klopf1972brain}
A~Harry Klopf,
\newblock {\em Brain function and adaptive systems: a heterostatic theory},
\newblock Number 133. Air Force Cambridge Research Laboratories, Air Force
  Systems Command, United~…, 1972.

\bibitem{hopfieldhebb}
Dmitry Krotov and John~J Hopfield,
\newblock ``Unsupervised learning by competing hidden units,''
\newblock {\em Proceedings of the National Academy of Sciences}, vol. 116, no.
  16, pp. 7723--7731, 2019.

\bibitem{HebbNet}
Manas Gupta, ArulMurugan Ambikapathi, and Savitha Ramasamy,
\newblock ``Hebbnet: A simplified hebbian learning framework to do biologically
  plausible learning,''
\newblock in {\em ICASSP 2021-2021 IEEE International Conference on Acoustics,
  Speech and Signal Processing (ICASSP)}. IEEE, 2021, pp. 3115--3119.

\bibitem{amato2019hebbian}
Giuseppe Amato, Fabio Carrara, Fabrizio Falchi, Claudio Gennaro, and Gabriele
  Lagani,
\newblock ``Hebbian learning meets deep convolutional neural networks,''
\newblock in {\em International Conference on Image Analysis and Processing}.
  Springer, 2019, pp. 324--334.

\bibitem{rumelhart1985feature}
David~E Rumelhart and David Zipser,
\newblock ``Feature discovery by competitive learning,''
\newblock {\em Cognitive science}, vol. 9, no. 1, pp. 75--112, 1985.

\bibitem{1992Neural}
C.~Gielen,
\newblock ``Neural computation and self-organizing maps, an introduction,''
\newblock {\em Neurocomputing}, vol. 5, no. 4-5, pp. 243--244, 1992.

\bibitem{supervisedSOM3}
Willem Melssen, Ron Wehrens, and Lutgarde Buydens,
\newblock ``Supervised kohonen networks for classification problems,''
\newblock {\em Chemometrics and Intelligent Laboratory Systems}, vol. 83, no.
  2, pp. 99--113, 2006.

\bibitem{dsom1}
Nan Liu, Jinjun Wang, and Yihong Gong,
\newblock ``Deep self-organizing map for visual classification,''
\newblock in {\em 2015 international joint conference on neural networks
  (IJCNN)}. IEEE, 2015, pp. 1--6.

\bibitem{dsom2}
Chathurika~S Wickramasinghe, Kasun Amarasinghe, and Milos Manic,
\newblock ``Deep self-organizing maps for unsupervised image classification,''
\newblock {\em IEEE Transactions on Industrial Informatics}, vol. 15, no. 11,
  pp. 5837--5845, 2019.

\bibitem{somcnn1}
Ehsan Mohebi and Adil Bagirov,
\newblock ``A convolutional recursive modified self organizing map for
  handwritten digits recognition,''
\newblock {\em Neural Networks}, vol. 60, pp. 104--118, 2014.

\bibitem{somcnn2}
Mohamed Sakkari and Mourad Zaied,
\newblock ``A convolutional deep self-organizing map feature extraction for
  machine learning,''
\newblock {\em Multimedia Tools and Applications}, vol. 79, no. 27, pp.
  19451--19470, 2020.

\bibitem{somcnn3}
Hiroshi Dozono, Gen Niina, and Satoru Araki,
\newblock ``Convolutional self organizing map,''
\newblock in {\em 2016 international conference on computational science and
  computational intelligence (CSCI)}. IEEE, 2016, pp. 767--771.

\bibitem{somcnn4}
Saleh Aly and Sultan Almotairi,
\newblock ``Deep convolutional self-organizing map network for robust
  handwritten digit recognition,''
\newblock {\em IEEE Access}, vol. 8, pp. 107035--107045, 2020.

\bibitem{golledge2003correlations}
Huw~DR Golledge, Stefano Panzeri, Fashan Zheng, Gianni Pola, Jack~W Scannell,
  Dimitrios~V Giannikopoulos, Roger~J Mason, Martin~J Tov{\'e}e, and Malcolm~P
  Young,
\newblock ``Correlations, feature-binding and population coding in primary
  visual cortex,''
\newblock {\em Neuroreport}, vol. 14, no. 7, pp. 1045--1050, 2003.

\bibitem{mnist}
Arthur Asuncion and David Newman,
\newblock ``Uci machine learning repository,'' 2007.

\bibitem{cifar10}
Alex Krizhevsky, Vinod Nair, and Geoffrey Hinton,
\newblock ``Cifar-10 (canadian institute for advanced research),''
\newblock {\em URL http://www. cs. toronto. edu/kriz/cifar. html}, vol. 5, no.
  4, pp. 1, 2010.

\bibitem{kandel2000principles}
Eric~R Kandel, James~H Schwartz, Thomas~M Jessell, Steven Siegelbaum, A~James
  Hudspeth, Sarah Mack, et~al.,
\newblock {\em Principles of neural science}, vol.~4,
\newblock McGraw-hill New York, 2000.

\bibitem{rolls1995sparseness}
Edmund~T Rolls and Martin~J Tovee,
\newblock ``Sparseness of the neuronal representation of stimuli in the primate
  temporal visual cortex,''
\newblock {\em Journal of neurophysiology}, vol. 73, no. 2, pp. 713--726, 1995.

\bibitem{rolls1997information}
Edmund~T Rolls, Alessandro Treves, Martin~J Tovee, and Stefano Panzeri,
\newblock ``Information in the neuronal representation of individual stimuli in
  the primate temporal visual cortex,''
\newblock {\em Journal of computational neuroscience}, vol. 4, no. 4, pp.
  309--333, 1997.

\bibitem{franco2007neuronal}
Leonardo Franco, Edmund~T Rolls, Nikolaos~C Aggelopoulos, and Jose~M Jerez,
\newblock ``Neuronal selectivity, population sparseness, and ergodicity in the
  inferior temporal visual cortex,''
\newblock {\em Biological cybernetics}, vol. 96, no. 6, pp. 547--560, 2007.

\end{thebibliography}

\end{document}